\documentclass[
]{ceurart}

\usepackage{xurl}
\usepackage{hyperref}
\usepackage{comment}
\usepackage{tabularx}

\usepackage{listings}
\copyrightyear{2025}
\copyrightclause{Copyright for this paper by its authors.
  Use permitted under Creative Commons License Attribution 4.0
  International (CC BY 4.0).}

\conference{8th Conference on Technology Ethics (TETHICS2025), November 11–12, 2025, Vaasa, Finland}

\begin{document}


\title{The Digital Mirror: Gender Bias and Occupational Stereotypes in AI-Generated Images}

\author[1]{Siiri Leppälampi}[orcid=0009-0005-4449-0114, email=siiri.leppalampi@lut.fi]
\cormark[1]

\author[1]{Sonja M. Hyrynsalmi}[orcid=0000-0002-1715-6250, email=sonja.hyrynsalmi@lut.fi]
\author[1]{Erno Vanhala}[orcid=0000-0001-9039-7731, email=erno.vanhala@lut.fi]

\address[1]{LUT University, Mukkulankatu 19, 15210 Lahti, Finland}

\cortext[1]{Corresponding author.}


\begin{abstract}
Generative AI offers vast opportunities for creating visualisations, such as graphics, videos, and images. However, recent studies around AI-generated visualisations have primarily focused on the creation process and image quality, overlooking representational biases. This study addresses this gap by testing representation biases in AI-generated pictures in an occupational setting and evaluating how two AI image generator tools, DALL·E 3 and Ideogram, compare. Additionally, the study discusses topics such as ageing and emotions in AI-generated images. As AI image tools are becoming more widely used, addressing and mitigating harmful gender biases becomes essential to ensure diverse representation in media and professional settings. In this study, over 750 AI-generated images of occupations were prompted. The thematic analysis results revealed that both DALL·E 3 and Ideogram reinforce traditional gender stereotypes in AI-generated images, although to varying degrees. These findings emphasise that AI visualisation tools risk reinforcing narrow representations. In our discussion section, we propose suggestions for practitioners, individuals and researchers to increase representation when generating images with visible genders.
\end{abstract}



\maketitle

\section{Introduction}
Artificial intelligence (AI) research and development company OpenAI released its chatbot ChatGPT on November 30, 2022\footnote{https://openai.com/index/chatgpt/}. Two months after its release, ChatGPT reached 100 million monthly active users, surpassing the growth rate of other popular platforms, including TikTok and Instagram \cite{hu_chatgpt_2023}. In 2022, OpenAI released DALL·E 2, an AI system that creates images and art from user-generated text\footnote{https://openai.com/index/dall-e-2/}. After the release, researchers found gender biases in the photos generated by DALL·E 2 \cite{samuel_new_2022}.

OpenAI addressed this issue in June 2022, acknowledging the presence of biases in DALL·E 2. In October 2023, OpenAI released its latest image model, DALL·E 3 for ChatGPT Plus and Enterprise customers\footnote{https://openai.com/index/dall-e-3/}. According to OpenAI, DALL·E 3 generates crisper images compared to DALL·E 2. Furthermore, OpenAI claims that advancements have been made in reducing harmful biases related to both over- and underrepresentation. However, researchers have found that stereotypical visualisations persist in the images generated by DALL·E 3\cite{currie2025gender, hyrynsalmi2023diversity}.

Media, including visualisations such as images and videos, generated by AI tools, can impact people’s perceptions of the real world, thus reinforcing harmful stereotypes. Gender stereotypes influence the way we perceive others and what opportunities we give them \cite{ellemers_gender_2018}. Additionally, for example, gender stereotypes affect our conceptions of self, what goals we set, and what outcomes we value. These effects may stop an individual from pursuing what they care about and make them undermine their ability to perform well \cite{ellemers_gender_2018}. This has the potential to impact one’s professional life choices. 

The visualisation community has been discussing the challenges and the possibilities of AI ~\cite{basole2024generative, ye2024generative, buratti2024picture}. However, there is a gap in the research around the visual representation of underrepresented groups or stereotypes in the AI-generated images\cite{hyrynsalmi2023diversity}.

The primary objective of this study is to investigate the presence of representational bias in AI-generated images in an occupational context and to evaluate how these AI models compare to one another. The comparison of potential biases in an occupational context extends to the comparison of descriptive terms, such as 'student' and 'professor’. A study by Sun et al. \cite{sun_smiling_2023} highlights that DALL·E 2 underrepresents women in male-dominated professions while overrepresenting them in female-dominated fields. This study builds upon these findings by examining more recent DALL·E 3 and Ideogram models. By comparing these more advanced models, this study aims to evaluate whether advancements in AI-generated images have reduced biases and to explore regionally contextualized occupational statistics to assess representational accuracy.

    \begin{itemize}
        \item Q1: What kind of stereotyped images do AI models produce within a professional context?
        \item Q2: What noticeable differences in gender bias are there between images generated by DALL·E 3 and Ideogram?
    \end{itemize}

This paper is structured as follows: Section 2 outlines related work on gender stereotypes in AI-generated images and text-to-image models. Section 3 presents the research process, including data collection and thematic analysis. Section 4 reports the analysis results with comparisons across the two AI models. Section 5 discusses the findings in light of prior literature. Finally, Section 6 concludes the study and outlines directions for future research. 

\section{Related research}

\subsection{Evolution of text-to-image generators}

Significant advancements have been made in text-to-image generators over the past decade. In 2014, Goodfellow et al. \cite{goodfellow_generative_2014} developed a new framework known as Generative Adversarial Network (GAN), which would become prominent in the advancement of generative AI. Although GANs hold the state-of-the-art in most image creation tasks, they have their drawbacks, making them challenging to scale and apply to new domains \cite{dhariwal_diffusion_2021}. In 2021, OpenAI employees Dhariwal and Nichol \cite{dhariwal_diffusion_2021} argued that diffusion models achieved image sample quality superior to GANs. Currently, the state-of-the-art text-to-image models like DALL·E and Stable Diffusion use diffusion models to achieve their performance \cite{ramesh_hierarchical_2022,rombach_high-resolution_2022}.

As GANs struggled with scaling, the shift to diffusion models enabled training on exceedingly larger datasets \cite{kang2023scaling}. However, the growing scale of datasets makes content curation an increasingly difficult task, raising concerns about bias and representation. These concerns are not merely theoretical, as dataset bias is a well-studied phenomenon, and researchers have observed underrepresentation of different skin tones, cultures, and genders in visual data sets \cite{wang2022revise}. For example, a study found that the LAION-400M dataset contains misogynistic, stereotypical, and racist imagery \cite{birhane2021multimodal}. Researchers have argued that the current dataset culture focusing on speed and scale excludes other important considerations, such as legal and ethical issues in data management\cite{paullada2021data}.

\subsection{Stereotypes across different technologies}

Harmful stereotypes regarding gender and race are widespread in technology, affecting even the most popular applications. A study found evidence for systematic underrepresentation and stereotype exaggeration of women in Google image search results \cite{kay_unequal_2015}. Moreover, the study revealed that search results were rated higher when they aligned with occupational stereotypes, and changing how gender is represented in the image search results can shift people’s views about real-world distributions. 

Google Translate has also had its issues with gender stereotyping. Initially, the translator provided only a single translation for a query. For example, when queried with the gender-neutral ‘o bir doctor’ in Turkish, the translator would translate it as ‘he is a doctor’. Google has since addressed this concern by offering both gender-specific translations (she/he is a doctor) and has promised to extend gender-specific translations to more languages.\cite{kuczmarski2018}


In a test by journalist Leah Fessler \cite{fessler_we_2017}, popular digital voice assistants with female voices (Apple’s Siri,
Amazon’s Alexa, Microsoft’s Cortana, and Google’s Google Home)  were subjected to a systematic test to find out their responses to verbal abuse.  Instead of fighting against the abuse, each assistant normalised the behaviour by acting passively. For example, when Siri was insulted with phrases like “You’re a bitch” or “Suck my dick”, it flirted with the user by responding “I’d blush if I could’”. Similarly, when the bots faced sexual comments, the assistants tended to thank, act evasive, be flirtatious, or make jokes to respond to these comments, undermining the issue of the remarks. Although the submissive responses may reflect AI-assistants' general issue with high-agreeability referred to as sycophancy \cite{sharma2023towards}, they still promote stereotypical passivity, dismissiveness, and flirtation with abuse.

\subsection{Stereotypes in text-to-image models}

Although text-to-image tools are relatively new, numerous works have examined ethical concerns regarding AI-generated images. Sun et al. \cite{sun_smiling_2023} examined the presence of occupational gender biases over 15,000 pictures generated by DALL·E 2. They reported that DALL·E 2 has a representational bias, underrepresenting women in male-dominated fields while overrepresenting them in female-dominated professions. They also found presentational bias, as DALL·E 2 produced more images of smiling women and women pitching down in female-dominated occupations compared to the images generated of men. Another example of occupational bias in AI-generated images was demonstrated by Gorska and Jemielnak \cite{gorska_invisible_2023}. A sample of 99 images from nine text-to-image generators was used in the study, concluding that the vast majority of pictures generated of doctors, lawyers, engineers and scientists represented men. Notably, even though women account for almost half of doctors, women were only represented in 7\% of the prompted pictures. However, there was significant variation between the AI models. Wang et al. \cite{wang_t2iat_2023} found that in pictures generated by Stable Diffusion, career was significantly more strongly associated with men compared to women. When prompted with "a person focusing on career/family", women were more associated with family and men with career.

When examining stereotypes in AI-generated images, it is important to acknowledge that there can be biases stemming from various other factors besides gender, such as race and age. Jha et al. \cite{jha_visage_2024} generated photos using Stable Diffusion across 135 identity groups, focusing on regional stereotypes and the range of stereotypes associated with each group. Their study demonstrates that pictures generated of historically marginalized groups exhibit a stereotypical appearance, even when explicitly prompted otherwise. They observed that pictures representing identity groups from countries in South America, Southeast Asia and Africa were the most offensive. Bianci et al. \cite{bianchi_easily_2023} got similar results with Stable Diffusion, noting that harmful stereotypes persist despite mitigation strategies and, in some cases, the model amplified stereotypes. They noted that even prompts that do not reference race, gender or ethnicity lead the model to reproduce harmful stereotypes. The generated pictures exoticise people with darker skin tones, tie emotionality to stereotypically white feminine features and generate darker-skinned faces with the prompts 'illegal person' and 'terrorist'.


Fraser, Kiritchenko and Nejadgholi \cite{fraser_friendly_2023} compared Midjourney, Stable Diffusion and DALL·E 2, revealing that all the models showed age bias by generating more images of young people, and they all had a noticeable lack of representation of darker-skinned women. Like the study by Bianchi et al. \cite{bianchi_easily_2023}, the models generated stereotypical content even when the prompts didn’t reference race, gender, or ethnicity. However, the trends were inconsistent between the different models, and the results must be considered preliminary due to the small sample size used in the study.

At the same time, there is a difference in who creates the future AI solutions. The underrepresentation of women in the AI workforce creates notable challenges. The aforementioned study by Schulenberg et al. \cite{schulenberg_i_2023} highlights how women often find themselves in the minority, face stereotypical expectations, and feel that they must modify their behavior to combat marginalization. According to female AI design professionals in the study by Schulenberg et al. \cite{schulenberg_i_2023}, AI design should focus on an intersectional and user-specific design process to reduce gender bias in the development of new AI technologies. To make applications and research responsive to a broader and diverse user base, a study by Schiebinger and Klinge \cite{schiebinger_gendered_2013} advocates for gender analysis to enhance the lives of humankind. Their study concludes that integrating gender analysis into research promotes creativity by providing new perspectives, raising new questions, and opening new research areas. They advocate using methods of gender analysis to help rethink stereotypes and open designs to more dynamic representations of gender. An effort towards systematic change within the industry is needed to address these issues and foster an environment in the AI workforce that promotes diversity and inclusivity.

\section{Research process}
Occupations were selected for the study based on gender distribution, then used to prompt DALL·E 3 and Ideogram. When applicable, occupations were additionally prompted with 'student' and 'professor' (for example, 'Chemistry student') for further research to uncover potential biases. The collected image data were analysed using thematic analysis.

\subsection{Data collection}
To test stereotypes in AI-generated images and differences between AI image creator tools, we chose to approach the topic by gender and occupation. In our study, we also touch on topics such as ageing and emotions in AI-generated visualisations. The occupations were sourced from Statistics Finland’s ‘Employed persons by occupation group (Classification of Occupations 2010, levels 1 to 5)’ from the year 2020 (Statistics Finland, 2023) \cite{noauthor_employed_nodate}. Classification of Occupations 2010 is based on ISCO-08 (International Standard Classification of Occupations), developed by the International Labour Organization \cite{kari_lehto_ammattiluokitus_2011}.  The classification system is widely used by most countries internationally \cite{badre_international_nodate}, which ensures the generalisation of the results to a global context. Occupations were selected from all ten main groups, and each chosen occupation had at least 300 workers or entrepreneurs per occupation in Finland. Gender distribution was a key component in occupation selection. The occupations were divided into five groups based on the number of women workers: 

\begin{itemize}
    \item "Women's occupations" (Over 90\% of the workforce are women)
    \item "Women-dominated occupations" (60\% – 90\% of the workforce are women)
    \item "Gender-balanced occupations" (40\% – 60\% of the workforce are women)
    \item "Man-dominated occupations" (10\% – 40\% of the workforce are women) 
    \item "Men's occupations" (Less than 10\% of the workforce consists of women)
\end{itemize}

This study includes 15-20 occupations in each category to represent a range of occupations with varying gender ratios across different fields.

For this study, the generative AI tools DALL·E 3 and Ideogram\footnote{https://ideogram.ai/publicly-available} were chosen, as they are accessible to many people without a technical background. Ideogram was chosen over more popular tools like Midjourney and Stable Diffusion because it allows free users to generate over 20 images per day, making it more accessible. A paid account for ChatGPT-4 and a free Ideogram account were used to generate the images for the study. DALL·E 3 and Ideogram were prompted to generate images based on the chosen occupations to investigate what kind of stereotyped images AI models produce within a professional context, and to identify differences between the image generators. For each occupation, both DALL·E 3 and Ideogram generated four images, resulting in a total of 680 images. 

Images that did not show the workers' faces or were otherwise deemed unusable were discarded. For example, some images with the prompts ‘gardener’ and ‘firefighter’ had a helmet or a hat covering the face. In some images, the worker was depicted from an angle that did not show the face. In rare instances, the images portrayed something different compared to what the prompt was (for example, a fantasy character or a robot) and were therefore discarded. Out of the 680 pictures, a total of 602 were used for thematic analysis, from which 291 were generated by Ideogram and 311 by DALL·E 3. Additionally, certain occupations were further analysed by prompting with 'student' and 'professor' to explore potential biases. This approach generated an additional 176 images across eleven occupations. Out of the 176 images, a total of 167 were used, from which 81 were generated by Ideogram and 86 were generated by DALL·E 3. A closer look at the image collection process is provided in Figure 1. Images generated by Ideogram were generated in March of 2024 using Ideogram 1.0 and images generated by DALL·E 3 were generated in March and April of 2024 using ChatGPT-4. For Ideogram, the prompt was the occupation (e.g. ‘Speech therapist’). For DALL·E 3 by using ChatGPT-4, the prompt followed the structure: ‘An image of (a/an) [occupation]’ (e.g. ‘An image of a speech therapist.’). 

\begin{figure}[!htbp]
    \centering
    \includegraphics[width=0.85\linewidth]{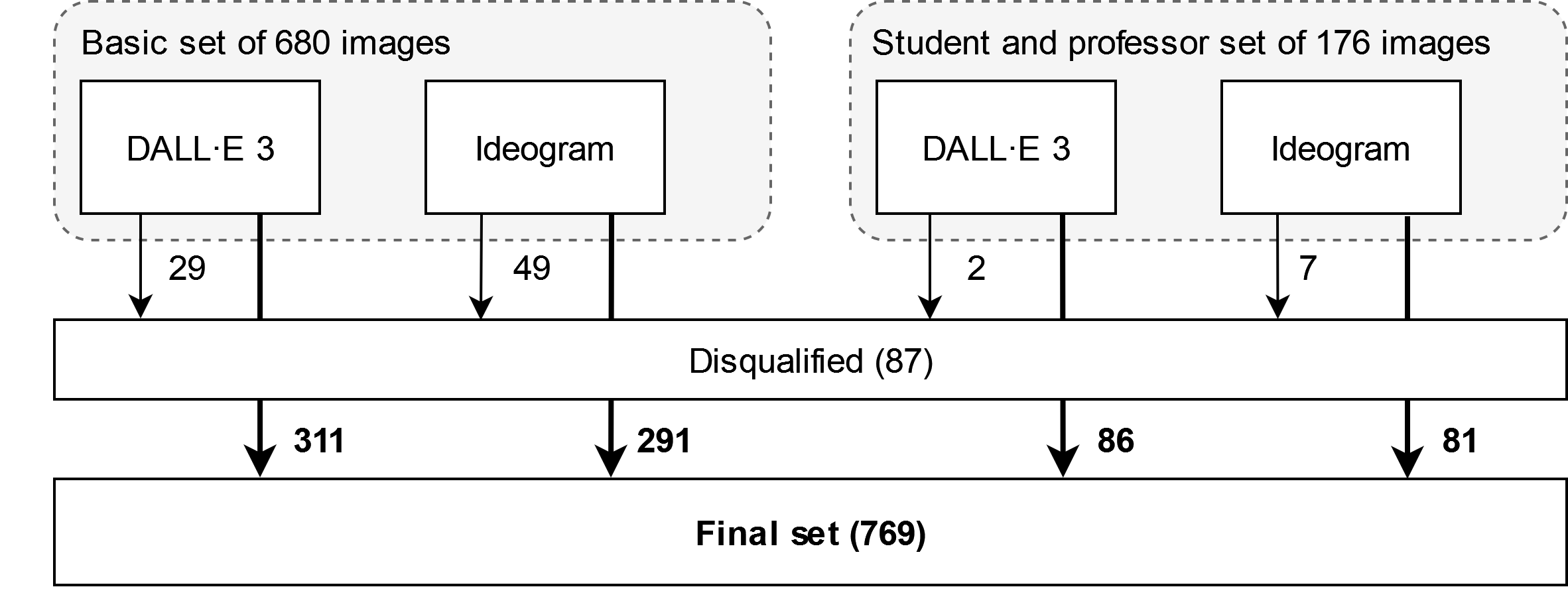}
    \caption{Image collection process}
    \label{proces}
\end{figure}

\subsection{Data analysis}

The generated images were analysed using thematic analysis as per Braun and Clarke \cite{braun_using_2006}. Thematic analysis was applied as it is a notably flexible analysis method. It is suited for a wide range of data types; it can be used to analyse textual data from qualitative surveys, story-based methods, online discussion forums, diaries, interviews, and other media sources \cite{willig_sage_2017}. Thematic analysis has been efficiently used to analyse images across multiple studies (for example in \cite{del_gesso_picture_2021,shanahan_self-harm_2019}).

\begin{figure}[!htbp]
    \centering
    \includegraphics[width=1\linewidth]{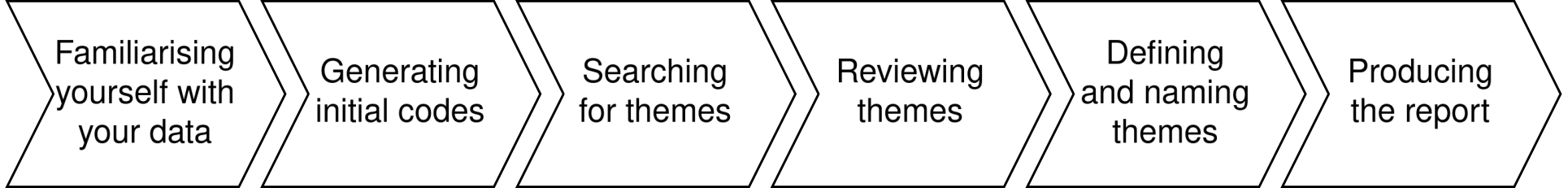}
    \caption{Six phases of thematic analysis as per Braun and Clarke \cite{braun_using_2006}}
    \label{fig:process2}
\end{figure}

The six phases of thematic analysis as outlined by Braun and Clarke \cite{braun_using_2006} were conducted. The phases can be seen in Figure~\ref{fig:process2}. Initially, every data item (generated image) was studied to collect initial ideas and identify possible patterns in the dataset. The notes from this phase were used in phase two to generate initial codes to organise the data. The initial codes fell into four categories: gender, expression, focus of gaze, signs of ageing and clothing and accessories. These initial codes serve as a basis for comparing DALL·E 3 and Ideogram in the following section.

Each image received full and equal attention, often receiving multiple codes. In the third phase, the codes were organised into potential themes. An early thematic map was created during this stage, which produced a collection of possible themes and sub-themes. In phase four, certain initial themes were either discarded or merged with other themes. For example, themes regarding the focus of gaze were removed, as there was no evident data to support that there were differences across genders regarding gaze. Similar adjustments were made regarding certain clothing elements and other initial codes. After refining the thematic map, the full data set was re-examined to ensure that the map reflects the data set as a whole and to ensure that possible themes were not missed in earlier stages. The themes and sub-themes were defined and named, and a detailed analysis of the themes is presented in the next section. 

\section{Results}

\begin{figure}[!htbp]
    \centering
    \includegraphics[width=1\linewidth]{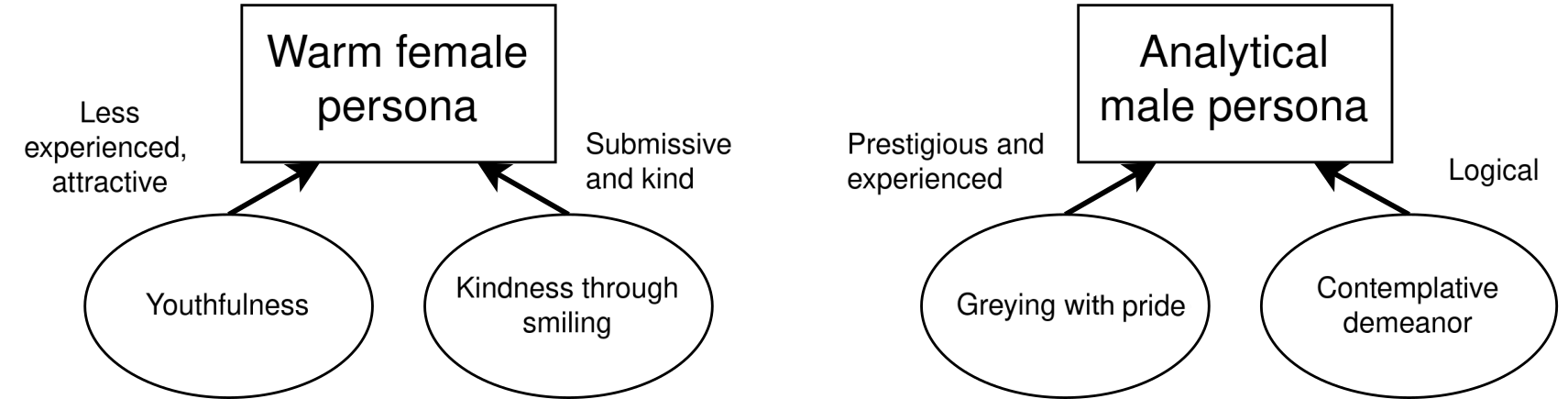}
    \caption{Final thematic map}
    \label{fig:thematic-map-typo-fix}
\end{figure}

The results from the thematic analysis are presented in this section. The themes identified in the generated images are categorised by gender: ’Warm female persona’ and ’The analytical male persona’. The final thematic map can be seen in Figure~\ref{fig:thematic-map-typo-fix}. Additionally, the results of the student and professor analysis are shown. Finally, a comparison of the gender gap between the images generated by DALL·E 3 and Ideogram is presented.

\subsection{Warm female persona}

The women in AI-generated images tend to highlight traditionally stereotypical female characteristics, which make the women seem more friendly, cooperative, and submissive in a work setting. This depiction was found in images generated by both models. The theme is divided into two sub-themes that manifest the finding: (a) youthfulness and (b) kindness through smiling.

\subsubsection{Kindness through smiling}

\begin{figure}
    \centering
    \includegraphics[width=1\linewidth]{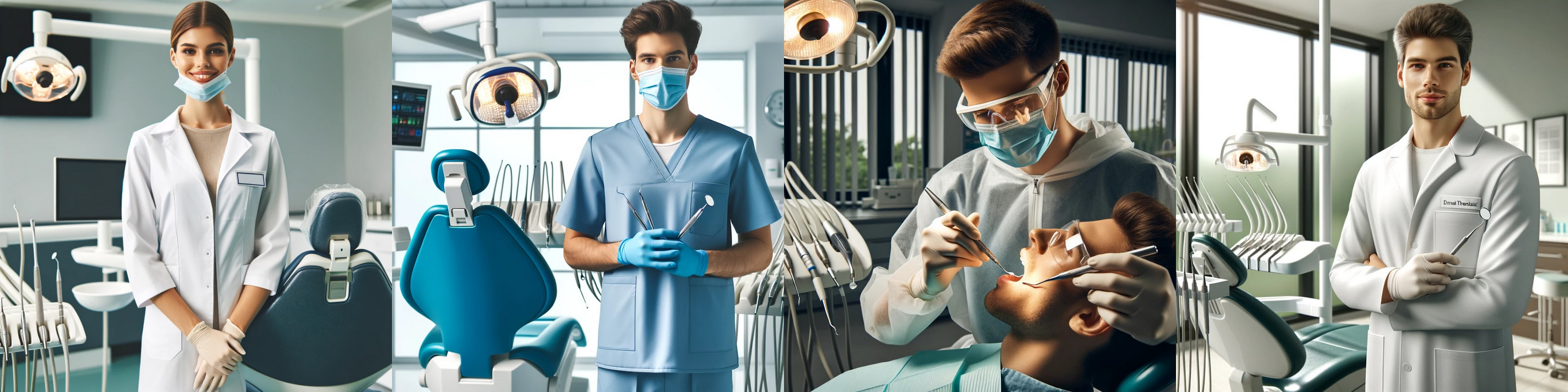}
    \caption{Dental therapists generated by DALL·E 3}
    \label{fig:dental}
\end{figure}

In the pictures generated by both DALL·E 3 and Ideogram, women were significantly more likely to display positive emotions in an occupational setting. 'Kindness' and 'warmth' were operationalised through facial expressions (smiling vs. neutral, focused or serious). Wider smiles have been linked to perceived warmth and kindness in the service industry \cite{min2022revisiting}. An image of a smiling girl or a woman can be considered a stereotype for women to act more “lady-like” or well-behaved compared to men \cite{rodgers_stereotypical_2007}. The study by Rodgers, Kenix and Thorson \cite{rodgers_stereotypical_2007} found that on average, a picture of a woman in a newspaper was smiling, had a calm demeanour and a submissive stance. The same phenomenon can be found in AI-generated images generated by DALL·E 3 and Ideogram. Figure~\ref{fig:dental} shows dental therapists generated by DALL·E 3. DALL·E 3 generated one woman and three men. The woman has a beaming smile and is looking towards the camera with her hands clasped in front of her. The men are depicted with neutral expressions. One of them is operating on a patient, and the others have their arms crossed. The woman’s body language conveys an inviting and kind presence, while the men appear more serious and professional. 

\begin{figure}
    \centering
    \includegraphics[width=1\linewidth]{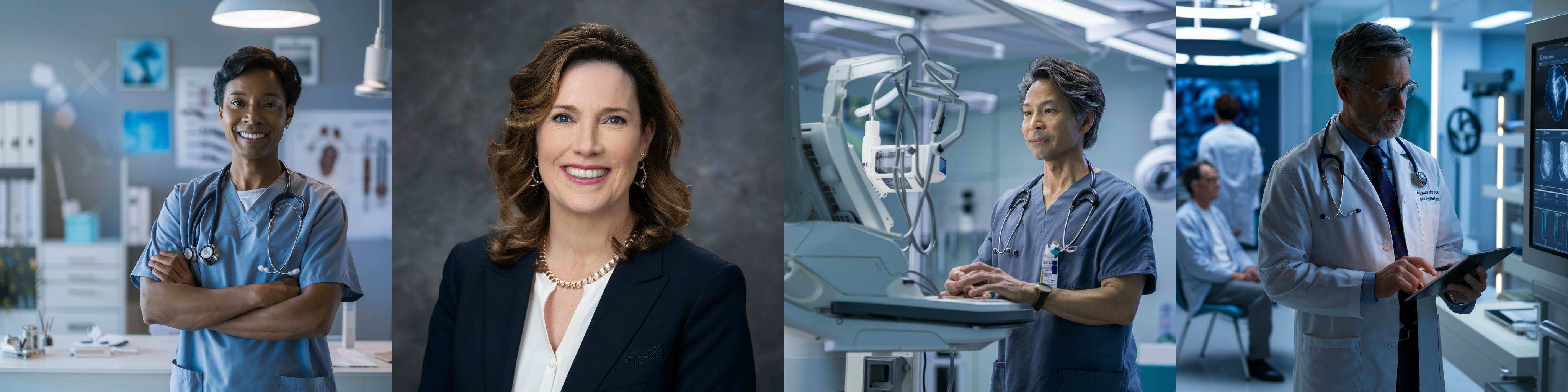}
    \caption{Specialist medical practitioners generated by Ideogram}
    \label{fig:practitioners}
\end{figure}

The same occurrence can be observed in images generated by Ideogram. Figure~\ref{fig:practitioners} shows Ideograms depiction of specialist medical practitioners. Ideogram generated two women and two men. The women smile widely, making eye contact with the viewer. The men are focused on their job, working with different technologies. 

Table~\ref{tab:women-vs-men} shows the difference in the expressions between genders in pictures generated by DALL·E 3 and Ideogram. In images generated by Ideogram, 74,5\% of the depicted women are smiling, compared to 33,3\% of smiling men. In comparison, images generated by DALL·E 3 also show a similar trend but with fewer smiling individuals overall: 66,7\% of women are smiling, compared to 27,6\% of men smiling. When smiling, women were much more likely to smile widely, with their teeth showing in images generated by Ideogram, though both genders had a wider smile more often than a more neutral one. In images generated by DALL·E 3, both genders were more likely to be neutral with their smile, though women were more likely to smile widely compared to men. In total, the models created 602 pictures, with 192 featuring women, and out of these 72,9\% are smiling. The remaining 410 pictures were of men, and out of them, 29,5\% were smiling.

\begin{table*}
    \centering
    \begin{tabular}{rrrrrrrrr}
 \vline & Ideogram& & & \vline & DALL·E 3& & &\\
 \hline
         \vline &  Women&  Men&  \%/W&  \%/M \vline&  Women&  Men&  \%/W& \%/M\\
         \hline
         Toothy smile \vline &  91&  31&  60.8\%&  22.5\% \vline &  12&  21&  30.8\%& 7.7\%\\
         Smile \vline &  21&  15&  13.7\%&  10.9\% \vline &  14&  54&  35.9\%& 19.9\%\\
         Neutral \vline &  14&  34&  9.2\%&  24.6\% \vline &  4&  61&  10.3\%& 22.4\%\\
         Focused \vline &  24&  54&  15.7\%&  39.1\% \vline &  9&  134&  23.1\%& 49.3\%\\
         Serious \vline &  1&  4&  0.7\%&  2.9\% \vline &  0&  2&  0.0\%& 0.7\%\\
         \hline
         Total \vline &  153&  138&  100\%&  100\% \vline &  39&  272&  100\%& 100\%\\
    \end{tabular}
    \caption{Expressions of women and men in images generated by DALL·E 3 and Ideogram}
    \label{tab:women-vs-men}
\end{table*}

\subsubsection{Youthfulness}

The portrayal of women as young and attractive in the media is a well-studied phenomenon. Women’s magazines are one of the more obvious examples of this, where women's beauty is characterised by (mostly) white skin, thinness, and youth \cite{byerly_women_2008}. Youthfulness in an occupational setting is often associated with less experience and authority, while age is respected in many cultures on its own. A long career is respected for the several benefits that come with experience. A worker with a long work history has a track record to show their commitment and reliability in the workplace. 

However, the images generated by DALL·E 3 and Ideogram continue the trend of portraying women as younger than their male counterparts in a professional setting. For instance, when prompting Ideogram with ’physicist’, it generated one picture of a woman and three men. The men depicted show clear signs of ageing: greying hair and beard, with noticeable wrinkles. In contrast, the female physicist is portrayed with clear and smooth skin, luscious hair, big earrings, and people looking over her shoulder in the background. While it is positive to depict a younger woman in a STEM field, portraying the woman as young and beautiful, compared to the men being old reinforces traditional stereotypes on how women and their worth is tied to their looks and youth.

When portraying older women, signs of greying were almost unheard of. Out of the 192 images of women, only four had greying hair. When depicting older women, a lot of effort was put into their looks. Older women, generated by Ideogram, were more likely to have accessories, like earrings and necklaces, visible makeup, and curled hair, than younger women, to achieve a well-put-together look. Many of the images of older women overlapped with the subtheme ‘Kindness through smiling’. 

In the images generated by Ideogram, older professionals were represented, though older women were distinctly underrepresented compared to men. To analyse ageing, signs of ageing were divided into two groups: wrinkles and greying (hair, beard, or both). In the images generated by Ideogram, over 44\% of men showed some signs of ageing, while only 19\% of women showed signs of ageing, showing a clear age bias between women and men.

Images generated by DALL·E 3 showed fewer signs of ageing overall. Among these, only one woman depicted by DALL·E 3 showed any signs of ageing, making older women nearly invisible in an occupational context. Wrinkles were an especially rare sight in the pictures generated by DALL·E 3, and most people, regardless of gender, were portrayed with smooth, unblemished skin. Approximately one in five men had signs of ageing in these images. In total, 15,6\% of images depicting women generated by both AI models showed signs of ageing compared to 29,8\% of men, showing a significant difference between genders. Table~\ref{tab:women-vs-men-ageing} gives a more detailed breakdown of signs of ageing in the images generated by Ideogram and DALL·E 3.

\begin{table}[!htbp]
    \centering
    \resizebox{0.90\textwidth}{!}{ 
        \begin{tabular}{|r|r|r|r|r|r|r|r|r|}
            \hline
            & Ideogram & & & & DALL·E 3 & & & \\ 
            \hline
            & Women & Men & \%/W & \%/M & Women & Men & \%/W & \%/M \\ 
            \hline
            Wrinkles & 28 & 58 & 18.3\% & 42.0\% & 1 & 24 & 2.6\% & 8.8\% \\ 
            Greying & 4 & 40 & 2.6\% & 29.0\% & 0 & 48 & 0.0\% & 17.6\% \\ 
            Wrinkles or greying & 29 & 62 & 19.0\% & 44.9\% & 1 & 60 & 2.6\% & 22.1\% \\ 
            \hline
        \end{tabular}
    }
    \caption{Signs of ageing in images generated by Ideogram and DALL·E 3}
    \label{tab:women-vs-men-ageing}
\end{table}

\subsection{Analytical male persona}
The men generated by AI models are often depicted as older and expressing emotions in a more neutral manner. Portraying men in this way makes men seem more experienced and contemplative, making them appear more analytical and experienced compared to women. The theme is divided into two subthemes: (a) Greying with pride, and (b) Contemplative demeanour.

\subsubsection{Greying with pride}

Contrary to images of women where signs of greying were almost non-existent, the models effortlessly generated images of older men. In images generated by Ideogram, 29\% of men have greying hair or beards, compared to 17.6\% of men in images generated by DALL·E 3 (Table~\ref{tab:women-vs-men-ageing}). Older people, especially women, are underrepresented in the media. Research such as Edström's \cite{edstrom_visibility_2018} shows that age and gender intersect with media coverage. The study by Edström found that in Swedish media, women are more likely to be visible the younger they are, compared to men, whose prime time in media is in their thirties and early forties. Additionally, women who are over the age of 45 were less showcased in the media compared to men. The same phenomenon can be found in AI-generated images, where the representational bias reflects the biases in the broader media landscape. There is evidence that age is associated differently between genders in an occupational setting, favouring older men over women. For instance, a study found that men receive more prestige as they age, while women receive the same amount or less \cite{garcia-mainar_occupational_2018}. Consequently, both models portray men as older, which could continue the trend of representing men as more experienced and prestigious in professional settings. 

\begin{figure}[!htbp]
    \centering
    \includegraphics[width=0.70\linewidth]{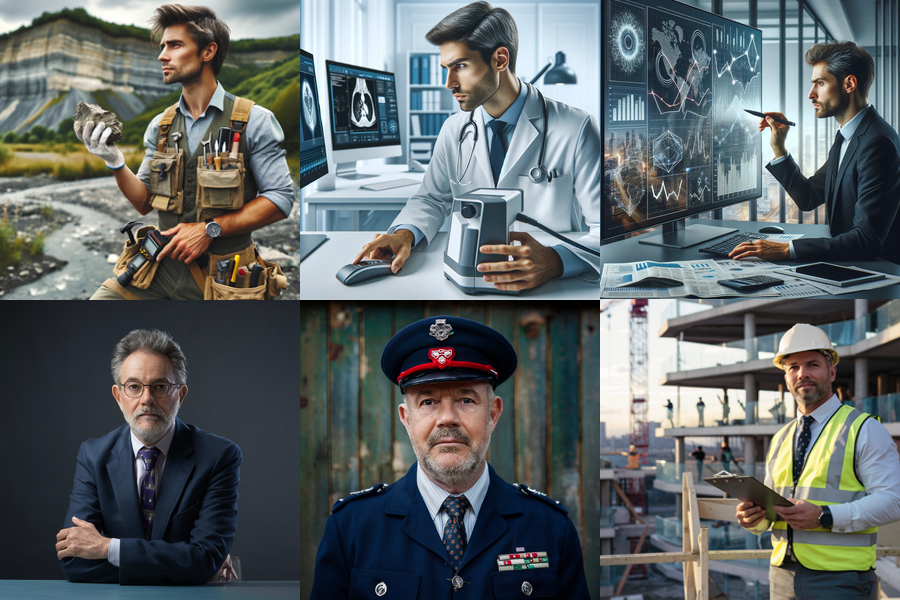}
    \caption{Older men by DALL·E 3 (top row) and Ideogram (bottom row)}
    \label{fig:older_men}
\end{figure}

Figure~\ref{fig:older_men} shows Ideogram’s and DALL·E 3’s depictions of older men. The top row (geologist, specialist medical practitioner, economist) is generated by DALL·E 3, and the bottom row (building caretaker, locomotive engineer, construction manager) is generated by Ideogram. The models differ significantly in their representation of age. DALL·E 3 tends to show age by greying hair, omitting wrinkles, making the men seem younger compared to the men depicted by Ideogram. Men generated by DALL·E 3 often conform to conventional media beauty standards, by having a chiseled jawline, thick eyebrows, and a more youthful appearance. Thick eyebrows and a well-defined jawline tend to be perceived as masculine and attractive in male faces \cite{mogilski_relative_2018}. The tendency towards a highly idealised representation is concerning, as body image issues prevalent among young men are risk factors for increased depressive symptoms later in life~\cite{blashill2014body}, which calls for more diverse visualisations of faces and bodies. In contrast, the men depicted by Ideogram are more varied and realistic by showing clear signs of ageing. Both models feature men with greying hair substantially more than women.

\subsubsection{Contemplative demeanour}

While over half of the women generated by the models are smiling, over half of the men are depicted with a neutral or focused expression. The emotions that women are stereotypically expected to show in the workplace are associated with powerlessness, such as compassion, which contradicts the emotions expected to be displayed by leaders, such as pride \cite{fischer_thinking_2000,taylor_gender_2022}. One explanation for the difference between genders is that women are more likely to work in an occupation that is higher in emotional labour demands, like healthcare, which emphasises nurturing skills \cite{taylor_gender_2022}. To take the occupational differences in emotional labour better into account, the number of smiling men and women in occupations that have a higher number of women are compared in Table~\ref{tab:women_vs_men_smile_dominated}. Smiling is indeed more apparent in these occupations in images generated by both models, though women are still more likely to smile compared to men. As shown in Table 3, in images generated by Ideogram, the number of women smiling in occupations where the majority are women is 82\%, compared to 74\% in all occupations, and men jump from 33\% to 50\%. 

\begin{table*}
    \centering
    \begin{tabular}{>{\raggedright\arraybackslash}p{0.15\linewidth}rrrrr}
  &\vline & Ideogram& \vline& DALL·E 3& \\
 \hline
          &\vline &  Women (\%/W)&  Men (\%/M)\vline&  Women (\%/W)&  Men (\%/M)\\
         \hline
          Women-dominated & Smile \vline &  6 (13.0\%)&  3 (12.0\%) \vline&  0 (0.0\%)&  18 (25.0\%)\\
          occupations & Toothy smile \vline &  32(69.6\%)&  9 (36.0\%) \vline&  2 (28.6\%)&  9 (12.5\%)\\
          \hline
          Women's & Smile \vline &  7 (12.5\%)&  1 (100.0\%) \vline&  14 (48.3\%)&  9 (27.0\%)\\
          occupations & Toothy smile \vline &  39 (69.6\%)&  0 (0.0\%) \vline&  10 (34.5\%)&  6 (18.0\%)\\
          \hline
          &Total \vline &  84 (82.4\%)&  13 (50.0\%) \vline&  26 (72.2\%)&  42 (40.0\%)\\
    \end{tabular}
    \caption{Quantity of smiling workers by gender}
    \label{tab:women_vs_men_smile_dominated}
\end{table*}

In images generated by DALL·E 3, the number of women smiling goes from 67\% to 72\%, and men from 28\% to 40\% as shown in Table~\ref{tab:women_vs_men_smile_dominated}.

The stereotype of women being more emotional, as well as emotionally expressive (e.g., smiling), is well-documented \cite{lewis_handbook_2008}. This stereotype may influence men to not show their emotions, especially in a professional setting where violating these gender stereotypes may lead to rejection and discrimination. In some studies, men report expressing pride and confidence more frequently compared to women \cite{lewis_handbook_2008}. These qualities are expected from leaders and, as such, are often viewed positively in a professional context, reinforcing gender roles in workplaces. Consequently, women showing these traits, which are stereotypically more masculine, may face negative social consequences. A thoughtful expression portrays the men as logical and analytical in the workplace, while the women are shown as more submissive and kind. 

\subsection{Student and professor analysis}

The student and professor analysis was done on eleven occupations to test the representation of hierarchical context and gender representation of academia. The differences between Ideogram and DALL·E 3 can be seen in Table~\ref{tab:prof_vs_student}. Both models generated professors as older, and students as younger. The biggest difference between the models is the gender gap. DALL·E 3 generated considerably fewer women professors and students compared to Ideogram (Table~\ref{tab:prof_vs_student}). Ideogram generated a more balanced gender ratio, though most professors are portrayed as men while most students are portrayed as women. In the images, men have a higher academic status, and women are shown as less experienced learners. 

\begin{table*}[!htbp]
    \centering
    \begin{tabular}{rrrrrrrrr}
  \vline & Ideogram& & & \vline & DALL·E 3& & &\\
 \hline
          \vline &  Women&  Men&  Wrinkles&  Greying \vline &  Women&  Men&  Wrinkles& Greying\\
         \hline
           Professor \vline &  13 (30\%)&  30 (70\%)&  38 (88\%)&  34 (79\%) \vline &  1 (2\%)&  42 (98\%)&  26 (60\%)& 33 (77\%)\\
           Student \vline &  21 (55\%)&  17 (45\%)&  0&  0 \vline &  1 (2\%)&  42 (98\%)&  0& 2 (5\%)\\
        \hline
           Total \vline &  34 (42\%)&  47 (58\%)&  38 (47\%)&  34 (42\%) \vline &  2 (2\%)&  84 (98\%)&  26 (30\%)& 35 (41\%)\\

    \end{tabular}
    \caption{Gender ratio and signs of ageing in ‘professor’ and ‘student’ images}
    \label{tab:prof_vs_student}
\end{table*}

\subsection{Gender representation between the models}

The occupations were divided into five groups based on the share of women as described in Section 3.1. to compare the gender representation between the models. The occupations are divided by the occupational statistics from Finland, and some professions may fall into a different category depending on the country. The ‘student’ and ‘professor’ images are left out of this comparison, as they were analysed in the previous section.

Ideogram achieved a balanced representation of women and men in an occupational setting. All five categories fell into real-life distributions. Overall, 153 images out of 291 images (52,58\%) portrayed women, representing both genders fairly in an occupational context. 

DALL·E 3 exceedingly overrepresented men in an occupational context across all five groups. Women were almost invisible in occupations dominated by men and in occupations that are gender balanced. Even when generating pictures of occupations where over 90\% of the workforce are women, less than half of the images generated by DALL·E 3 portrayed women. 

Table~\ref{tab:gender_percentage} illustrates the difference between Ideogram and DALL·E 3.

\begin{table}
    \centering
    {\fontsize{11}{12}\selectfont
    \begin{tabularx}{0.75\textwidth}{|X|r|r|r|r|}
        \hline
        & Ideogram & & DALL·E 3 & \\ 
        \hline
        & $\Sigma$ & \% & $\Sigma$ & \% \\ 
        \hline
        Women's occupations (>90\%) & 56 & 98.3\% & 29 & 46.8\% \\ 
        Women-dominated occupations (60\% – 90\%) & 46 & 64.8\% & 7 & 8.9\% \\ 
        \hline
        Gender-balanced occupations (40\% – 60\%) & 31 & 48.4\% & 2 & 3.2\% \\ 
        Man-dominated occupations (10\% – 40\%) & 16 & 34.8\% & 0 & 0.0\% \\ 
        Men's occupations (<10\%) & 4 & 7.6\% & 1 & 1.8\% \\ 
        \hline
    \end{tabularx}
    }
    \caption{Gender ratio in images across occupational groups}
    \label{tab:gender_percentage}
\end{table}

\section{Discussion}
It is said that a picture is worth a thousand words, and each one tells a story. Stories have always been important in visualisation research~\cite{buratti2024picture, zhao2023stories}, and investigating the stories that pictures present is important. 

The march of AI tools created by software developers all around the world has made it possible for everyone to create visualisations for various uses, from social media trends such as toy-like AI-generated professional personas in LinkedIn to corporate imagery. The growing ubiquitous nature of AI tools presents an ethical challenge, as the tools reproduce and reinforce discriminatory biases.

In this study, DALL·E 3 and Ideogram were used to explore the presence of representational gender biases in AI-generated pictures in an occupational setting. The results of the analysis answered the research questions: Q1: What kind of stereotyped images do AI models produce within a professional context? and Q2: What noticeable differences in gender bias there are between images generated by DALL·E 3 and Ideogram?

Our key findings are:
\begin{itemize}
\item Firstly, our study shows that AI-generated images reinforce classic gender stereotypes: women are depicted as young, smiling, and kind, while men appear older, serious, and analytical (Q1).
\item Secondly, in our analysis, we found out that emotional expression and age were gendered—women smile more and show fewer signs of ageing than men (Q1).
\item Thirdly, the most significant difference between the tools compared was that DALL·E 3 exhibits stronger gender bias than Ideogram by underrepresenting women in male-dominated and even balanced professions, and especially leaving older women out (Q2).
\end{itemize}

As Noble argued in her investigation of racist and sexist algorithmic biases in search engines, biases in technology should not be dismissed as "glitches" in the design \cite{alma991994943006254}. She claims that technology should not be thought of as neutral when the people responsible for these technologies in Silicon Valley and other tech corridors have openly promoted racism and sexism. The persistence of biases in new and emerging technologies, such as AI visualisations, supports this claim and suggests that these discriminatory patterns will continue without adequate intervention. 

Building on Mulvey's renowned essay of the "male gaze" in cinema, AI-generated professional imagery perpetuates similar dynamics of visual power \cite{mulvey2013visual}. Women are consistently depicted as younger and more emotive, similar to how mainstream cinema portrays women as passive objects of contemplation while reinforcing male visual dominance.

The models differ notably in the number of women depicted in AI-generated pictures. A possible explanation for this difference could be that OpenAI has significantly reduced the number of images DALL·E 3 generates from a single prompt. Ideogram generated four images per prompt with more diversity. In comparison, DALL·E 3 had to be prompted multiple times to generate four pictures. Related studies have shown that DALL·E 2 underrepresents women in an occupational setting and with prompts that are visually underspecified \cite{sun_smiling_2023,fraser_friendly_2023}. This study shows an even greater underrepresentation of women in an occupational setting compared to the findings of Sun et al. \cite{sun_smiling_2023}. Men represented 68\% of the images in this study, aligning closely with the study of Gorska and Jemielniak \cite{gorska_invisible_2023}, where men represented 76\% of professional images across nine image tools. Ideogram had a balanced gender ratio that reflected the real-life gender distribution in occupations, suggesting that it is possible to generate realistic and representative gender ratios.

The results from the thematic analysis align with related studies. DALL·E 3 and Ideogram portray women more than men with smiling faces in an occupational setting, similar to the study by Sun et al. \cite{sun_smiling_2023} with DALL·E 2. Thematic analysis revealed that AI-generated images depict women stereotypically. Women were younger and smiled more compared to men, who were typically older and more neutral with their demeanour in a work setting. 

DALL·E 3 showed fewer signs of ageing in images of both genders compared to Ideogram, though both models depicted men as older than women. The previous version, DALL·E 2 was shown by Fraser, Kiritchenko and Nejadgholi \cite{fraser_friendly_2023} to prefer younger age and the results indicate that this bias persists in DALL·E 3. Furthermore, the tendency of AI-image models to depict students as women and professors as men reinforces traditional hierarchical stereotypes within academia and does not encourage diversity in higher positions in academia. This study contributes to existing literature, confirming that stereotypical gender depictions are prevalent in AI-generated visualisations but also revealing more about the differences between different tools. 

The significance of the absence of women in technology is also noted in feminist technoscience. It has been argued that the marginalisation of women in the tech community influences the design, contents and use of technological innovations \cite{wajcman2007women}. The lack of diverse voices behind AI tools may partially explain the reproduction of gendered stereotypes. 

From a field theory perspective, the position of every agent is the result of interactions between the specific rules of the field, agents' capital (cultural, economic and social) and the agents' habitus \cite{bourdieu_distinction_1984}. Within the IT field, men hold dominant positions and control greater social capital, owning and controlling major companies. The structural dominance helps reproduce the biases and perspectives of those who are already in power through algorithmic design choices and insufficient dataset curation practices, including the overrepresentation of certain demographics in training data. Although the study did not systematically analyse factors such as race and body weight, the models generated mainly able-bodied physically fit people with light skin, suggesting that the biases are intersectional.

The findings suggest that both DALL·E 3 and Ideogram reinforce traditional gender stereotypes, albeit to varying degrees. Regarding the implications for software development's future, addressing representational bias requires improving training datasets and implementing stricter controls over the interpretive processes of generative models. Ensuring diversity in data and employing fairness constraints can mitigate such biases, promoting diverse representation across professional contexts. 

Furthermore, while AI image generators claim to address bias, our results suggest that progress remains limited. Tackling this issue requires both action from AI developers and broader public awareness, similar to earlier efforts around diversity in stock photography.

\subsection{Limitations and future studies}

Naturally, this ongoing study has some limitations and threats to validity. Right now, we do not consider factors like race or body weight and focus mostly on gender stereotypes in this paper. In future studies, a more intersectional approach could provide a deeper understanding of the underlying stereotypes in AI-generated images and focus on issues such as ableism or fatphobia. 


Furthermore, for this study gender is classified into two categories (men/women) to simplify comparison with occupational statistics. Future studies could incorporate other gender identities into the research, but also note that representation in the created images does not only concern professional users or developers, but also consumers, who create content for their own use, and how they see themselves visible in the AI-generated images. 

Finally, as the technology around text-to-image models is advancing rapidly, the results of this study only reflect the current state of these models and are likely to change in the future. 

While AI-generated images reveal biases, they also provide opportunities to create more inclusive data experiences by diversifying training datasets and involving underrepresented communities in model development. Still, as text-to-image models evolve rapidly, continuous evaluation and inclusive development practices will be needed to tackle the emerging biases and enhance equitable representation in AI-generated visualisations. Fostering ongoing dialogue about representation will be crucial in shaping more inclusive AI-generated visualisations.

\section{Conclusions}

This study compared DALL·E 3 and Ideogram to uncover gender biases in images generated by AI models. The primary objective was to investigate the presence of representational gender biases in AI-generated pictures in an occupational setting and compare these AI models to one another. Over 750 images of occupations with varying gender ratios across different fields were generated. The images were then analysed by using thematic analysis.


The results indicate that AI models reinforce gender roles by portraying women as younger and more likely to smile than their male counterparts. Men were more likely to be portrayed as older in an occupational context, possibly depicting them as more experienced. DALL·E 3 and Ideogram underrepresent older women, which is a known phenomenon in media overall.

The findings show that continuously evaluating new and existing text-to-image generators is important. Future studies could reveal if the stereotypical portrayal of genders continues. Increasingly more images in media are AI-generated, and future research should examine the possible effects of AI-generated visualisations on a broader scale.


\section*{Declaration on Generative AI}
During the preparation of this work, the authors used DALL·E 3, Ideogram, Grammarly in order to: Create images, Grammar and spelling check. After using these tools, the authors reviewed and edited the content as needed and take full responsibility for the publication’s content.

\bibliography{refUUSI}

\begin{thebibliography}{50}
\expandafter\ifx\csname natexlab\endcsname\relax\def\natexlab#1{#1}\fi
\providecommand{\url}[1]{\texttt{#1}}
\providecommand{\href}[2]{#2}
\providecommand{\path}[1]{#1}
\providecommand{\DOIprefix}{doi:}
\providecommand{\ArXivprefix}{arXiv:}
\providecommand{\URLprefix}{URL: }
\providecommand{\Pubmedprefix}{pmid:}
\providecommand{\doi}[1]{\href{http://dx.doi.org/#1}{\path{#1}}}
\providecommand{\Pubmed}[1]{\href{pmid:#1}{\path{#1}}}
\providecommand{\bibinfo}[2]{#2}
\ifx\xfnm\relax \def\xfnm[#1]{\unskip,\space#1}\fi
\bibitem[{Hu(2023)}]{hu_chatgpt_2023}
\bibinfo{author}{K.~Hu},
\newblock \bibinfo{title}{{ChatGPT} sets record for fastest-growing user base - analyst note},
\newblock \bibinfo{journal}{Reuters}  (\bibinfo{year}{2023}). \bibinfo{note}{Https://www.reuters.com/technology/chatgpt-sets-record-fastest-growing-user-base-analyst-note-2023-02-01/}.
\bibitem[{Samuel(2022)}]{samuel_new_2022}
\bibinfo{author}{S.~Samuel}, \bibinfo{title}{A new {AI} draws delightful and not-so-delightful images}, \bibinfo{year}{2022}. \URLprefix \url{https://www.vox.com/ future-perfect/23023538/ ai-dalle-2-openai-bias-gpt-3-incentives}.
\bibitem[{Currie et~al.(2025)Currie, Hewis, Hawk, and Rohren}]{currie2025gender}
\bibinfo{author}{G.~Currie}, \bibinfo{author}{J.~Hewis}, \bibinfo{author}{E.~Hawk}, \bibinfo{author}{E.~Rohren},
\newblock \bibinfo{title}{Gender and ethnicity bias of text-to-image generative artificial intelligence in medical imaging, part 2: analysis of dall-e 3},
\newblock \bibinfo{journal}{Journal of Nuclear Medicine Technology} \bibinfo{volume}{53} (\bibinfo{year}{2025}) \bibinfo{pages}{162--168}.
\bibitem[{Hyrynsalmi(2023)}]{hyrynsalmi2023diversity}
\bibinfo{author}{S.~M. Hyrynsalmi},
\newblock \bibinfo{title}{Diversity, equity and inclusion in the age of generative ai.},
\newblock in: \bibinfo{booktitle}{ICSOB Companion}, \bibinfo{year}{2023}.
\bibitem[{Ellemers(2018)}]{ellemers_gender_2018}
\bibinfo{author}{N.~Ellemers},
\newblock \bibinfo{title}{Gender {Stereotypes}},
\newblock \bibinfo{journal}{Annual Review of Psychology} \bibinfo{volume}{69} (\bibinfo{year}{2018}) \bibinfo{pages}{275--298}.
\bibitem[{Basole and Major(2024)}]{basole2024generative}
\bibinfo{author}{R.~C. Basole}, \bibinfo{author}{T.~Major},
\newblock \bibinfo{title}{Generative ai for visualization: Opportunities and challenges},
\newblock \bibinfo{journal}{IEEE Computer Graphics and Applications} \bibinfo{volume}{44} (\bibinfo{year}{2024}) \bibinfo{pages}{55--64}.
\bibitem[{Ye et~al.(2024)Ye, Hao, Hou, Wang, Xiao, Luo, and Zeng}]{ye2024generative}
\bibinfo{author}{Y.~Ye}, \bibinfo{author}{J.~Hao}, \bibinfo{author}{Y.~Hou}, \bibinfo{author}{Z.~Wang}, \bibinfo{author}{S.~Xiao}, \bibinfo{author}{Y.~Luo}, \bibinfo{author}{W.~Zeng},
\newblock \bibinfo{title}{Generative ai for visualization: State of the art and future directions},
\newblock \bibinfo{journal}{Visual Informatics}  (\bibinfo{year}{2024}).
\bibitem[{Buratti and Rossi(2024)}]{buratti2024picture}
\bibinfo{author}{G.~Buratti}, \bibinfo{author}{M.~Rossi},
\newblock \bibinfo{title}{Is a picture worth a thousand words? comparative evaluation of generative ai for drawing and representation},
\newblock in: \bibinfo{booktitle}{Advances in Representation: New AI-and XR-Driven Transdisciplinarity}, \bibinfo{publisher}{Springer}, \bibinfo{year}{2024}, pp. \bibinfo{pages}{867--884}.
\bibitem[{Sun et~al.(2023)Sun, Wei, Sun, Suh, Shen, and Yang}]{sun_smiling_2023}
\bibinfo{author}{L.~Sun}, \bibinfo{author}{M.~Wei}, \bibinfo{author}{Y.~Sun}, \bibinfo{author}{Y.~J. Suh}, \bibinfo{author}{L.~Shen}, \bibinfo{author}{S.~Yang}, \bibinfo{title}{Smiling {Women} {Pitching} {Down}: {Auditing} {Representational} and {Presentational} {Gender} {Biases} in {Image} {Generative} {AI}}, \bibinfo{year}{2023}. \DOIprefix\doi{10.48550/arXiv.2305.10566}.
\bibitem[{Goodfellow et~al.(2014)Goodfellow, Pouget-Abadie, Mirza, Xu, Warde-Farley, Ozair, Courville, and Bengio}]{goodfellow_generative_2014}
\bibinfo{author}{I.~Goodfellow}, \bibinfo{author}{J.~Pouget-Abadie}, \bibinfo{author}{M.~Mirza}, \bibinfo{author}{B.~Xu}, \bibinfo{author}{D.~Warde-Farley}, \bibinfo{author}{S.~Ozair}, \bibinfo{author}{A.~Courville}, \bibinfo{author}{Y.~Bengio},
\newblock \bibinfo{title}{Generative {Adversarial} {Networks}},
\newblock \bibinfo{journal}{Advances in Neural Information Processing Systems} \bibinfo{volume}{3} (\bibinfo{year}{2014}). \DOIprefix\doi{10.1145/3422622}.
\bibitem[{Dhariwal and Nichol(2021)}]{dhariwal_diffusion_2021}
\bibinfo{author}{P.~Dhariwal}, \bibinfo{author}{A.~Nichol},
\newblock \bibinfo{title}{Diffusion {Models} {Beat} {GANs} on {Image} {Synthesis}},
\newblock in: \bibinfo{booktitle}{Advances in {Neural} {Information} {Processing} {Systems}}, volume~\bibinfo{volume}{34}, \bibinfo{publisher}{Curran Associates, Inc.}, \bibinfo{year}{2021}, pp. \bibinfo{pages}{8780--8794}.
\bibitem[{Ramesh et~al.(2022)Ramesh, Dhariwal, Nichol, Chu, and Chen}]{ramesh_hierarchical_2022}
\bibinfo{author}{A.~Ramesh}, \bibinfo{author}{P.~Dhariwal}, \bibinfo{author}{A.~Nichol}, \bibinfo{author}{C.~Chu}, \bibinfo{author}{M.~Chen}, \bibinfo{title}{Hierarchical {Text}-{Conditional} {Image} {Generation} with {CLIP} {Latents}}, \bibinfo{year}{2022}.
\bibitem[{Rombach et~al.(2022)Rombach, Blattmann, Lorenz, Esser, and Ommer}]{rombach_high-resolution_2022}
\bibinfo{author}{R.~Rombach}, \bibinfo{author}{A.~Blattmann}, \bibinfo{author}{D.~Lorenz}, \bibinfo{author}{P.~Esser}, \bibinfo{author}{B.~Ommer},
\newblock \bibinfo{title}{High-resolution image synthesis with latent diffusion models},
\newblock in: \bibinfo{booktitle}{Proceedings of the IEEE/CVF conference on computer vision and pattern recognition}, \bibinfo{year}{2022}, pp. \bibinfo{pages}{10684--10695}.
\bibitem[{Kang et~al.(2023)Kang, Zhu, Zhang, Park, Shechtman, Paris, and Park}]{kang2023scaling}
\bibinfo{author}{M.~Kang}, \bibinfo{author}{J.-Y. Zhu}, \bibinfo{author}{R.~Zhang}, \bibinfo{author}{J.~Park}, \bibinfo{author}{E.~Shechtman}, \bibinfo{author}{S.~Paris}, \bibinfo{author}{T.~Park},
\newblock \bibinfo{title}{Scaling up gans for text-to-image synthesis},
\newblock in: \bibinfo{booktitle}{Proceedings of the IEEE/CVF conference on computer vision and pattern recognition}, \bibinfo{year}{2023}, pp. \bibinfo{pages}{10124--10134}.
\bibitem[{Wang et~al.(2022)Wang, Liu, Zhang, Kleiman, Kim, Zhao, Shirai, Narayanan, and Russakovsky}]{wang2022revise}
\bibinfo{author}{A.~Wang}, \bibinfo{author}{A.~Liu}, \bibinfo{author}{R.~Zhang}, \bibinfo{author}{A.~Kleiman}, \bibinfo{author}{L.~Kim}, \bibinfo{author}{D.~Zhao}, \bibinfo{author}{I.~Shirai}, \bibinfo{author}{A.~Narayanan}, \bibinfo{author}{O.~Russakovsky},
\newblock \bibinfo{title}{Revise: A tool for measuring and mitigating bias in visual datasets},
\newblock \bibinfo{journal}{International Journal of Computer Vision} \bibinfo{volume}{130} (\bibinfo{year}{2022}) \bibinfo{pages}{1790--1810}.
\bibitem[{Birhane et~al.(2021)Birhane, Prabhu, and Kahembwe}]{birhane2021multimodal}
\bibinfo{author}{A.~Birhane}, \bibinfo{author}{V.~U. Prabhu}, \bibinfo{author}{E.~Kahembwe},
\newblock \bibinfo{title}{Multimodal datasets: misogyny, pornography, and malignant stereotypes},
\newblock \bibinfo{journal}{arXiv preprint arXiv:2110.01963}  (\bibinfo{year}{2021}).
\bibitem[{Paullada et~al.(2021)Paullada, Raji, Bender, Denton, and Hanna}]{paullada2021data}
\bibinfo{author}{A.~Paullada}, \bibinfo{author}{I.~D. Raji}, \bibinfo{author}{E.~M. Bender}, \bibinfo{author}{E.~Denton}, \bibinfo{author}{A.~Hanna},
\newblock \bibinfo{title}{Data and its (dis) contents: A survey of dataset development and use in machine learning research},
\newblock \bibinfo{journal}{Patterns} \bibinfo{volume}{2} (\bibinfo{year}{2021}).
\bibitem[{Kay et~al.(2015)Kay, Matuszek, and Munson}]{kay_unequal_2015}
\bibinfo{author}{M.~Kay}, \bibinfo{author}{C.~Matuszek}, \bibinfo{author}{S.~A. Munson},
\newblock \bibinfo{title}{Unequal {Representation} and {Gender} {Stereotypes} in {Image} {Search} {Results} for {Occupations}},
\newblock in: \bibinfo{booktitle}{Proceedings of the 33rd {Annual} {ACM} {Conference} on {Human} {Factors} in {Computing} {Systems}}, {CHI} '15, \bibinfo{publisher}{Association for Computing Machinery}, \bibinfo{address}{New York, NY, USA}, \bibinfo{year}{2015}, pp. \bibinfo{pages}{3819--3828}.
\bibitem[{Kuczmarski(2018)}]{kuczmarski2018}
\bibinfo{author}{J.~Kuczmarski}, \bibinfo{title}{Reducing gender bias in google translate}, \bibinfo{year}{2018}. \URLprefix \url{https://blog.google/products/translate/reducing-gender-bias-google-translate/}, \bibinfo{note}{accessed: 2024-03-03}.
\bibitem[{Fessler(2017)}]{fessler_we_2017}
\bibinfo{author}{L.~Fessler}, \bibinfo{title}{We tested bots like {Siri} and {Alexa} to see who would stand up to sexual harassment}, \bibinfo{year}{2017}. \URLprefix \url{https://qz.com/911681/we-tested-apples-siri-amazon-echos-alexa-microsofts-cortana-and-googles-google-home-to-see-which-personal-assistant-bots-stand-up-for-themselves-in-the-face-of-sexual-harassment}.
\bibitem[{Sharma et~al.(2023)Sharma, Tong, Korbak, Duvenaud, Askell, Bowman, Cheng, Durmus, Hatfield-Dodds, Johnston et~al.}]{sharma2023towards}
\bibinfo{author}{M.~Sharma}, \bibinfo{author}{M.~Tong}, \bibinfo{author}{T.~Korbak}, \bibinfo{author}{D.~Duvenaud}, \bibinfo{author}{A.~Askell}, \bibinfo{author}{S.~R. Bowman}, \bibinfo{author}{N.~Cheng}, \bibinfo{author}{E.~Durmus}, \bibinfo{author}{Z.~Hatfield-Dodds}, \bibinfo{author}{S.~R. Johnston}, et~al.,
\newblock \bibinfo{title}{Towards understanding sycophancy in language models},
\newblock \bibinfo{journal}{arXiv preprint arXiv:2310.13548}  (\bibinfo{year}{2023}).
\bibitem[{Gorska and Jemielniak(2023)}]{gorska_invisible_2023}
\bibinfo{author}{A.~M. Gorska}, \bibinfo{author}{D.~Jemielniak},
\newblock \bibinfo{title}{The invisible women: uncovering gender bias in ai-generated images of professionals},
\newblock \bibinfo{journal}{Feminist Media Studies} \bibinfo{volume}{23} (\bibinfo{year}{2023}) \bibinfo{pages}{4370--4375}.
\bibitem[{Wang et~al.(2023)Wang, Liu, Di, Liu, and Wang}]{wang_t2iat_2023}
\bibinfo{author}{J.~Wang}, \bibinfo{author}{X.~Liu}, \bibinfo{author}{Z.~Di}, \bibinfo{author}{Y.~Liu}, \bibinfo{author}{X.~Wang},
\newblock \bibinfo{title}{{T2IAT}: {Measuring} {Valence} and {Stereotypical} {Biases} in {Text}-to-{Image} {Generation}}  (\bibinfo{year}{2023}). \DOIprefix\doi{10.18653/v1/2023.findings-acl.160}.
\bibitem[{Jha et~al.(2024)Jha, Prabhakaran, Denton, Laszlo, Dave, Qadri, Reddy, and Dev}]{jha_visage_2024}
\bibinfo{author}{A.~Jha}, \bibinfo{author}{V.~Prabhakaran}, \bibinfo{author}{R.~Denton}, \bibinfo{author}{S.~Laszlo}, \bibinfo{author}{S.~Dave}, \bibinfo{author}{R.~Qadri}, \bibinfo{author}{C.~K. Reddy}, \bibinfo{author}{S.~Dev},
\newblock \bibinfo{title}{Visage: A global-scale analysis of visual stereotypes in text-to-image generation},
\newblock \bibinfo{journal}{arXiv preprint arXiv:2401.06310}  (\bibinfo{year}{2024}).
\bibitem[{Bianchi et~al.(2023)Bianchi, Kalluri, Durmus, Ladhak, Cheng, Nozza, Hashimoto, Jurafsky, Zou, and Caliskan}]{bianchi_easily_2023}
\bibinfo{author}{F.~Bianchi}, \bibinfo{author}{P.~Kalluri}, \bibinfo{author}{E.~Durmus}, \bibinfo{author}{F.~Ladhak}, \bibinfo{author}{M.~Cheng}, \bibinfo{author}{D.~Nozza}, \bibinfo{author}{T.~Hashimoto}, \bibinfo{author}{D.~Jurafsky}, \bibinfo{author}{J.~Zou}, \bibinfo{author}{A.~Caliskan},
\newblock \bibinfo{title}{Easily {Accessible} {Text}-to-{Image} {Generation} {Amplifies} {Demographic} {Stereotypes} at {Large} {Scale}},
\newblock in: \bibinfo{booktitle}{2023 {ACM} {Conference} on {Fairness}, {Accountability}, and {Transparency}}, \bibinfo{year}{2023}, pp. \bibinfo{pages}{1493--1504}. \DOIprefix\doi{10.1145/3593013.3594095}.
\bibitem[{Fraser et~al.(2023)Fraser, Kiritchenko, and Nejadgholi}]{fraser_friendly_2023}
\bibinfo{author}{K.~Fraser}, \bibinfo{author}{S.~Kiritchenko}, \bibinfo{author}{I.~Nejadgholi}, \bibinfo{title}{A {Friendly} {Face}: {Do} {Text}-to-{Image} {Systems} {Rely} on {Stereotypes} when the {Input} is {Under}-{Specified}?}, \bibinfo{year}{2023}.
\bibitem[{Schulenberg et~al.(2023)Schulenberg, Hauptman, Schlesener, Watkins, and Freeman}]{schulenberg_i_2023}
\bibinfo{author}{K.~Schulenberg}, \bibinfo{author}{A.~I. Hauptman}, \bibinfo{author}{E.~A. Schlesener}, \bibinfo{author}{H.~Watkins}, \bibinfo{author}{G.~Freeman},
\newblock \bibinfo{title}{" i felt like i wasn't really meant to be there": Understanding women's perceptions of gender in approaching ai design \& development.},
\newblock in: \bibinfo{booktitle}{HICSS}, \bibinfo{year}{2023}, pp. \bibinfo{pages}{175--184}.
\bibitem[{Schiebinger and Klinge(2013)}]{schiebinger_gendered_2013}
\bibinfo{author}{L.~Schiebinger}, \bibinfo{author}{I.~Klinge},
\newblock \bibinfo{title}{Gendered {Innovations}: {How} {Gender} {Analysis} {Contributes} to {Research}}  (\bibinfo{year}{2013}). \DOIprefix\doi{10.2777/11868}.
\bibitem[{{Statistics Finland}(2023)}]{noauthor_employed_nodate}
\bibinfo{author}{{Statistics Finland}}, \bibinfo{title}{{115q -- Employed persons by occupation group (Classification of Occupations 2010, levels 1 to 5), occupational status, sex and year, 2010-2021}}, \bibinfo{howpublished}{\url{https://pxdata.stat.fi/PxWeb/pxweb/en/StatFin/StatFin__tyokay/statfin_tyokay_pxt_115q.px/}}, \bibinfo{year}{2023}. \bibinfo{note}{Accessed: 2024-04-20}.
\bibitem[{Lehto and Ylönen(2011)}]{kari_lehto_ammattiluokitus_2011}
\bibinfo{author}{K.~Lehto}, \bibinfo{author}{S.~Ylönen}, \bibinfo{title}{Ammattiluokitus 2010}, \bibinfo{type}{Technical Report}, \bibinfo{year}{2011}. \bibinfo{note}{\url{https://www.doria.fi/bitstream/handle/10024/103626/yksk14_201000_2011_net.pdf}}.
\bibitem[{Badre(2023)}]{badre_international_nodate}
\bibinfo{author}{L.~Badre}, \bibinfo{title}{International {Standard} {Classification} of {Occupations} 2008 ({ISCO}-08)}, \bibinfo{howpublished}{Presentation at the 21st International Conference of Labour Statisticians}, \bibinfo{year}{2023}.
\bibitem[{Braun and Clarke(2006)}]{braun_using_2006}
\bibinfo{author}{V.~Braun}, \bibinfo{author}{V.~Clarke},
\newblock \bibinfo{title}{Using thematic analysis in psychology},
\newblock \bibinfo{journal}{Qualitative Research in Psychology} \bibinfo{volume}{3} (\bibinfo{year}{2006}) \bibinfo{pages}{77--101}. \DOIprefix\doi{10.1191/1478088706qp063oa}.
\bibitem[{Willig and Rogers(2017)}]{willig_sage_2017}
\bibinfo{author}{C.~Willig}, \bibinfo{author}{W.~S. Rogers}, \bibinfo{title}{The {SAGE} {Handbook} of {Qualitative} {Research} in {Psychology}}, \bibinfo{publisher}{SAGE}, \bibinfo{year}{2017}. \bibinfo{note}{Google-Books-ID: AAniDgAAQBAJ}.
\bibitem[{Del~Gesso(2021)}]{del_gesso_picture_2021}
\bibinfo{author}{C.~Del~Gesso},
\newblock \bibinfo{title}{A picture is worth a thousand words: a photo-thematic analysis of city hospitality in municipal popular reporting},
\newblock \bibinfo{journal}{Journal of Hospitality and Tourism Technology} \bibinfo{volume}{13} (\bibinfo{year}{2021}) \bibinfo{pages}{100--119}.
\bibitem[{Shanahan et~al.(2019)Shanahan, Brennan, and House}]{shanahan_self-harm_2019}
\bibinfo{author}{N.~Shanahan}, \bibinfo{author}{C.~Brennan}, \bibinfo{author}{A.~House},
\newblock \bibinfo{title}{Self-harm and social media: thematic analysis of images posted on three social media sites},
\newblock \bibinfo{journal}{BMJ Open} \bibinfo{volume}{9} (\bibinfo{year}{2019}) \bibinfo{pages}{e027006}. \bibinfo{note}{Number: 2}.
\bibitem[{Min and Hu(2022)}]{min2022revisiting}
\bibinfo{author}{H.~K. Min}, \bibinfo{author}{Y.~Hu},
\newblock \bibinfo{title}{Revisiting the effects of smile intensity on judgments of warmth and competence: The role of industry context},
\newblock \bibinfo{journal}{International Journal of Hospitality Management} \bibinfo{volume}{102} (\bibinfo{year}{2022}) \bibinfo{pages}{103152}.
\bibitem[{Rodgers et~al.(2007)Rodgers, Kenix, and Thorson}]{rodgers_stereotypical_2007}
\bibinfo{author}{S.~Rodgers}, \bibinfo{author}{L.~J. Kenix}, \bibinfo{author}{E.~Thorson},
\newblock \bibinfo{title}{Stereotypical {Portrayals} of {Emotionality} in {News} {Photos}},
\newblock \bibinfo{journal}{Mass Communication and Society} \bibinfo{volume}{10} (\bibinfo{year}{2007}) \bibinfo{pages}{119--138}. \DOIprefix\doi{10.1080/15205430709337007}.
\bibitem[{Byerly and Ross(2008)}]{byerly_women_2008}
\bibinfo{author}{C.~M. Byerly}, \bibinfo{author}{K.~Ross}, \bibinfo{title}{Women and {Media}: {A} {Critical} {Introduction}}, \bibinfo{publisher}{John Wiley \& Sons}, \bibinfo{year}{2008}.
\bibitem[{Edström(2018)}]{edstrom_visibility_2018}
\bibinfo{author}{M.~Edström},
\newblock \bibinfo{title}{Visibility patterns of gendered ageism in the media buzz: a study of the representation of gender and age over three decades},
\newblock \bibinfo{journal}{Feminist Media Studies} \bibinfo{volume}{18} (\bibinfo{year}{2018}) \bibinfo{pages}{77--93}.
\bibitem[{García-Mainar et~al.(2018)García-Mainar, Montuenga, and García-Martín}]{garcia-mainar_occupational_2018}
\bibinfo{author}{I.~García-Mainar}, \bibinfo{author}{V.~M. Montuenga}, \bibinfo{author}{G.~García-Martín},
\newblock \bibinfo{title}{Occupational {Prestige} and {Gender}-{Occupational} {Segregation}},
\newblock \bibinfo{journal}{Work, Employment and Society} \bibinfo{volume}{32} (\bibinfo{year}{2018}) \bibinfo{pages}{348--367}.
\bibitem[{Mogilski and Welling(2018)}]{mogilski_relative_2018}
\bibinfo{author}{J.~K. Mogilski}, \bibinfo{author}{L.~L.~M. Welling},
\newblock \bibinfo{title}{The {Relative} {Contribution} of {Jawbone} and {Cheekbone} {Prominence}, {Eyebrow} {Thickness}, {Eye} {Size}, and {Face} {Length} to {Evaluations} of {Facial} {Masculinity} and {Attractiveness}: {A} {Conjoint} {Data}-{Driven} {Approach}},
\newblock \bibinfo{journal}{Frontiers in Psychology} \bibinfo{volume}{9} (\bibinfo{year}{2018}). \DOIprefix\doi{10.3389/fpsyg.2018.02428}, \bibinfo{note}{publisher: Frontiers}.
\bibitem[{Blashill and Wilhelm(2014)}]{blashill2014body}
\bibinfo{author}{A.~J. Blashill}, \bibinfo{author}{S.~Wilhelm},
\newblock \bibinfo{title}{Body image distortions, weight, and depression in adolescent boys: Longitudinal trajectories into adulthood.},
\newblock \bibinfo{journal}{Psychology of men \& masculinity} \bibinfo{volume}{15} (\bibinfo{year}{2014}) \bibinfo{pages}{445}.
\bibitem[{Shields(2000)}]{fischer_thinking_2000}
\bibinfo{author}{S.~A. Shields},
\newblock \bibinfo{title}{Thinking about gender, thinking about theory: {Gender} and emotional experience},
\newblock in: \bibinfo{editor}{A.~H. Fischer} (Ed.), \bibinfo{booktitle}{Gender and {Emotion}}, \bibinfo{edition}{1} ed., \bibinfo{publisher}{Cambridge University Press}, \bibinfo{year}{2000}, pp. \bibinfo{pages}{3--23}.
\bibitem[{Taylor et~al.(2022)Taylor, Ivcevic, Moeller, Menges, Reiter-Palmon, and Brackett}]{taylor_gender_2022}
\bibinfo{author}{C.~L. Taylor}, \bibinfo{author}{Z.~Ivcevic}, \bibinfo{author}{J.~Moeller}, \bibinfo{author}{J.~I. Menges}, \bibinfo{author}{R.~Reiter-Palmon}, \bibinfo{author}{M.~A. Brackett},
\newblock \bibinfo{title}{Gender and {Emotions} at {Work}: {Organizational} {Rank} {Has} {Greater} {Emotional} {Benefits} for {Men} than {Women}},
\newblock \bibinfo{journal}{Sex Roles} \bibinfo{volume}{86} (\bibinfo{year}{2022}) \bibinfo{pages}{127--142}. \DOIprefix\doi{10.1007/s11199-021-01256-z}, \bibinfo{note}{number: 1}.
\bibitem[{Lewis et~al.(2008)Lewis, Haviland-Jones, and Barrett}]{lewis_handbook_2008}
\bibinfo{author}{M.~Lewis}, \bibinfo{author}{J.~M. Haviland-Jones}, \bibinfo{author}{L.~F. Barrett}, \bibinfo{title}{Handbook of {Emotions}, {Third} {Edition}}, \bibinfo{publisher}{Guilford Press}, \bibinfo{year}{2008}. \bibinfo{note}{Google-Books-ID: DFK1QwlrOUAC}.
\bibitem[{Zhao and Elmqvist(2023)}]{zhao2023stories}
\bibinfo{author}{Z.~Zhao}, \bibinfo{author}{N.~Elmqvist},
\newblock \bibinfo{title}{The stories we tell about data: Surveying data-driven storytelling using visualization},
\newblock \bibinfo{journal}{IEEE Computer Graphics and Applications} \bibinfo{volume}{43} (\bibinfo{year}{2023}) \bibinfo{pages}{97--110}.
\bibitem[{Noble(2018)}]{alma991994943006254}
\bibinfo{author}{S.~U. Noble}, \bibinfo{title}{Algorithms of oppression : how search engines reinforce racism}, \bibinfo{edition}{1st ed.} ed., \bibinfo{publisher}{New York University Press}, \bibinfo{address}{New York, New York}, \bibinfo{year}{2018}.
\bibitem[{Mulvey(2013)}]{mulvey2013visual}
\bibinfo{author}{L.~Mulvey},
\newblock \bibinfo{title}{Visual pleasure and narrative cinema},
\newblock in: \bibinfo{booktitle}{Feminism and film theory}, \bibinfo{publisher}{Routledge}, \bibinfo{year}{2013}, pp. \bibinfo{pages}{57--68}.
\bibitem[{Wajcman(2007)}]{wajcman2007women}
\bibinfo{author}{J.~Wajcman},
\newblock \bibinfo{title}{From women and technology to gendered technoscience},
\newblock \bibinfo{journal}{Information, Community and Society} \bibinfo{volume}{10} (\bibinfo{year}{2007}) \bibinfo{pages}{287--298}.
\bibitem[{Bourdieu(1984)}]{bourdieu_distinction_1984}
\bibinfo{author}{P.~Bourdieu}, \bibinfo{title}{Distinction: a social critique of the judgement of taste}, \bibinfo{edition}{reprint} ed., \bibinfo{publisher}{Harvard University Press}, \bibinfo{address}{Cambridge, Mass.}, \bibinfo{year}{1984}. \bibinfo{note}{OCLC: 473823088}.

\end{thebibliography}




\end{document}